\documentclass{esannV2}
\usepackage{enumitem}
\usepackage{graphicx,url,hyperref}
\usepackage{xcolor,wrapfig,algorithmic,subfig}
\usepackage{tabularx}
  \newcolumntype{C}[1]{>{\centering\arraybackslash}p{#1}}
  \newcolumntype{R}[1]{>{\raggedleft\arraybackslash}p{#1}}
  \newcolumntype{L}[1]{>{\raggedright\arraybackslash}p{#1}}
\usepackage{booktabs} 
\usepackage[ruled,linesnumbered]{algorithm2e}
\usepackage{amssymb}
\usepackage{amsmath}
\usepackage{mathtools}
\usepackage{xspace}
\usepackage{listings}
\usepackage{floatrow}

\newfloatcommand{capbtabbox}{figure}[][\FBwidth]
\usepackage{tikz}

\begin{document}

\date{}

\title{Unlocking Structured Thinking in Language Models with Cognitive Prompting}

%\title{LLaMA Tunes CMA-ES: Optimizing Performance with a Large Language Model}

%\title{Tuning CMA-ES Hyperparameters with LLaMA}

%LLaMA Tuning CMA-ES
%Tuning CMA-ES with LLaMA
%LLaMA for Tuning CMA-ES

\author{Oliver Kramer and Jill Baumann
%
% Optional short acknowledgment: remove next line if non-needed
%\thanks{This is an optional funding source acknowledgement.}
%
% DO NOT MODIFY THE FOLLOWING '\vspace' ARGUMENT
\vspace{.3cm}\\
%
% Addresses and institutions (remove "1- " in case of a single institution)
%Computational Intelligence Lab\\
Department of Computing Science\\Carl von Ossietzky Universität Oldenburg \\
26111 Oldenburg, Germany\\
\url{{oliver.kramer,jill.baumann}@uni-oldenburg.de}
%\texttt{\{oliver.kramer,jill.baumann\}@uni-oldenburg.de}
%
% Remove the next three lines in case of a single institution
}

\maketitle
\thispagestyle{empty}

\begin{abstract}
We propose cognitive prompting as a novel approach to guide problem-solving in large language models (LLMs) through structured, human-like cognitive operations, such as goal clarification, decomposition, filtering, abstraction, and pattern recognition. By employing systematic, step-by-step reasoning, cognitive prompting enables LLMs to tackle complex, multi-step tasks more efficiently. We introduce three variants: a deterministic sequence of cognitive operations, a self-adaptive variant in which the LLM dynamically selects the sequence of cognitive operations, and a hybrid variant that uses generated correct solutions as few-shot chain-of-thought prompts. Experiments with LLaMA, Gemma~2, and Qwen models in each two sizes on the arithmetic reasoning benchmark GSM8K demonstrate that cognitive prompting significantly improves performance compared to standard question answering.
\end{abstract}

\section{Introduction}

Recent advancements in AI, particularly in LLMs, have significantly improved tasks such as text summarization, code generation, and question answering. However, LLMs still face challenges with multi-step reasoning compared to human cognition.

This paper introduces cognitive prompting (CP), a method designed to enhance LLM problem-solving by emulating human cognitive operations (COPs) through structured steps such as goal clarification, decomposition, and pattern recognition (see Figure \ref{fig:cp}). Inspired by cognitive psychology, CP aims to bridge the gap between human reasoning and AI, improving performance in domains such as mathematics, logic, and decision-making.
Our experiments with LLaMA, Gemma 2, and Qwen models, each in two different sizes, on the GSM8K dataset \cite{gsm8k}, demonstrate significant performance gains, particularly with the hybrid of self-adaptive and few-shot chain-of-thought (CoT) variant.

The structure of the paper is as follows: Section \ref{sec:related} reviews related work; Section \ref{sec:cp} introduces the concept of CP;  Section \ref{sec:variants} describes three CP variants; Section \ref{sec:arith} presents experimental results on the impact of CP on arithmetic reasoning tasks; and Section \ref{sec:cons} concludes the paper.

\section{Related Work}
\label{sec:related}

Zero-shot prompting generates responses without providing specific examples, while few-shot prompting \cite{brown2020language} improves performance by including task-specific examples. CoT prompting \cite{wei2022chain} further enhances reasoning by breaking complex problems into sequential steps, enabling the model to process each stage independently. Tree of Thoughts (ToT) prompting \cite{tree} expands this approach by exploring multiple reasoning paths simultaneously, making it well-suited for intricate decision-making scenarios. ReAct \cite{yao2022react} integrates logical reasoning with real-time decision-making, offering enhanced adaptability in dynamic and interactive environments.
Prompt Breeder \cite{promptbreeder2022} employs evolutionary computation to iteratively optimize prompts for improved results. 
%Self-consistency \cite{wang2023self} reduces response variability by selecting the most frequent or consistent output from multiple generations.
Automated Prompt Engineering (APE) \cite{ape} and Optimization by PROmpting (OPRO) \cite{opro} take prompting refinement further by automating the design process. These methods often outperform manually crafted prompts by leveraging optimization algorithms to fine-tune instructions for optimal model performance.

\section{Cognitive Prompting}
\label{sec:cp}

CP structures problem-solving into a sequence of COPs, enabling LLMs to address complex tasks across domains like mathematics, logic, and decision-making. Drawing from cognitive psychology, CP breaks problems into stages that mimic human task refinement, enhancing clarity, interpretability, and adaptability. Unlike methods such as CoT \cite{wei2022chain}, CP provides multi-dimensional depth without manual solution design.

\begin{figure}[h!]
\centering
\includegraphics[width=\textwidth]{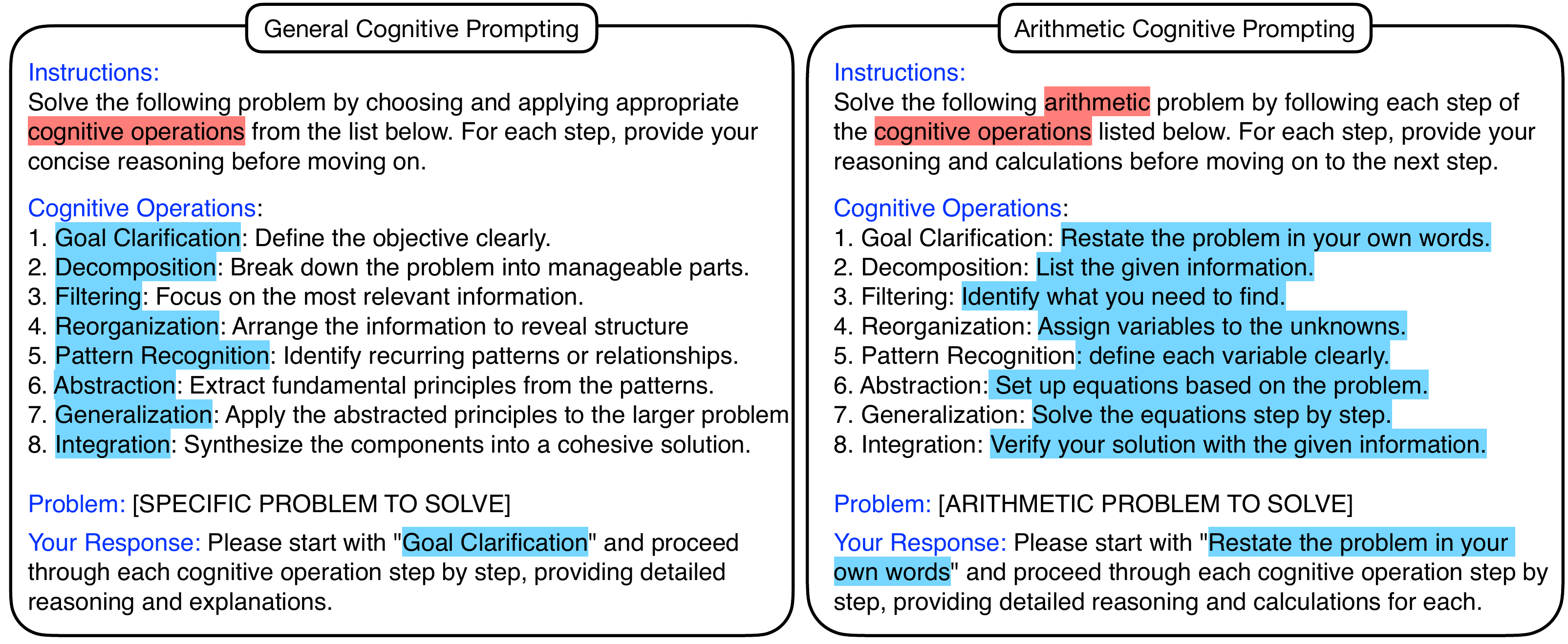}
\caption{\label{fig:cp}Left: General CP, Right: CP adapted to arithmetical reasoning.} 
\end{figure}

CP can be formalized as an optimization problem. Given a set of COPs \( C = \{c_1, c_2, \dots, c_n\} \) and a sequence \( S = \{s_1, s_2, \dots, s_k\} \) of \( k \) operations from \( C \), the goal is to find \( S^* \) that maximizes task performance \(
S^* = \arg \max_{S \subseteq C} f(S)
\)
subject to constraints like \( |S| = k \), \( s_1 = \text{goal clarification} \), and \( s_k = \text{integration} \). Here, \( f(S) \) measures performance (e.g., accuracy or coherence).

\paragraph{Cognitive Operations.} This paper focuses on eight key COPs.

\begin{itemize}[leftmargin=0pt]

    \item \textit{Goal Clarification.} This operation aligns the model’s reasoning with the desired outcome and minimizes distractions. All subsequent operations are guided by this goal.
    \item \textit{Decomposition:} Break the problem \( P \) into smaller sub-problems, \( P_1, P_2, \dots, P_n \). This incremental approach is particularly useful for complex, multi-step problems, such as mathematical proofs or logical reasoning. Decomposition isolates critical components for systematic problem-solving.

    \item \textit{Filtering:} Select the most relevant information from the problem set, \( I_{\text{rel}} \subseteq I \). Filtering ensures the model concentrates on key details, excluding irrelevant data. By narrowing its focus, the model achieves greater accuracy and efficiency in problem-solving.

    \item \textit{Reorganization:} Rearrange data or variables to reveal patterns or simplify the problem structure. Reorganization helps the model uncover underlying relationships, making complex data more interpretable, and is particularly effective for algebraic manipulation or logical structuring.

    \item \textit{Pattern Recognition:} Identify recurring patterns or relationships, \( \mathcal{P} \), that connect the problem to known solutions. Recognizing patterns accelerates problem-solving by allowing the model to apply established strategies. This enhances predictive accuracy and facilitates generalization.

    \item \textit{Abstraction:} Extract broader principles from the identified patterns, \( \mathcal{P} \), for application across different problems. Abstraction helps the model transcend specific details and focus on core concepts, enabling flexible problem-solving.

    \item \textit{Generalization:} Apply the abstracted principles to solve broader problems or similar contexts. Generalization ensures that solutions are scalable and adaptable to related tasks, enhancing the model's reasoning robustness and versatility.

    \item \textit{Integration:} Synthesize the individual solutions, \( Q_i \), into a cohesive final answer, \( Q \), ensuring all sub-problems are resolved and producing a unified and consistent solution.
\end{itemize}

\paragraph{Domain-specific COPs.}
Adapting COPs to specific domains ensures that the reasoning process remains relevant and effective for each task. For arithmetic reasoning, these general COPs are adapted as follows (see Figure \ref{fig:cp}, right).

\section{Cognitive Prompting Variants}
\label{sec:variants}
CP comes in three variants.
\textit{Deterministic cognitive prompting (D-CP)} follows a fixed manual designed sequence of cognitive operations, providing structure but less adaptability. We optimized the sequence of COPs in preliminary experiments. \textit{Self-adaptive cognitive prompting (SA-CP)} allows the model to self-select the next COP based on the task’s needs, i.e., the LLM decides on its own, which COP to choose next. A prompt incorporating the following command enables self-adaptive prompting:
\begin{lstlisting}
For each step, choose and apply the most suitable cognitive operation 
from the list below and provide a concise explanation of your reasoning 
before moving on to the next step.
\end{lstlisting}
This flexibility enhances problem-solving and produces more interpretable reasoning, but is based on the model's own ability to structure reasoning. \textit{Hybrid cognitive prompting (H-CP)} uses a brief LLM-generated summary of successful problem solutions previously generated with CP and adds all $k$ summaries to the CP instruction in a few-shot CoT fashion. This variant is based on the idea two combine structured thinking with successfully solved examples, a problem-solving strategy we believe also human reasoning often follows.

\section{Arithmetic Reasoning}
\label{sec:arith}

\paragraph{Benchmark.}
We evaluate the performance of CP using Meta’s LLaMA 3.1 (8B and 70B), Google’s Gemma 2 (9B and 27B), and Alibaba's Qwen 2.5 (7B and 32B) models on the GSM8K dataset \cite{gsm8k}, a widely used benchmark for math problem-solving. GSM8K consists of 7,000 training and 1,500 test high-quality, grade-school math word problems, designed to assess the reasoning and mathematical capabilities of LLMs. Since CP does not require training, we exclusively evaluate performance on the test set.

\paragraph{Mid-Size Models.}

\begin{figure}[h!]
    \centering
    \subfloat[LLaMA 8B\label{fig:llama_8b}]{
        \includegraphics[width=0.31\textwidth]{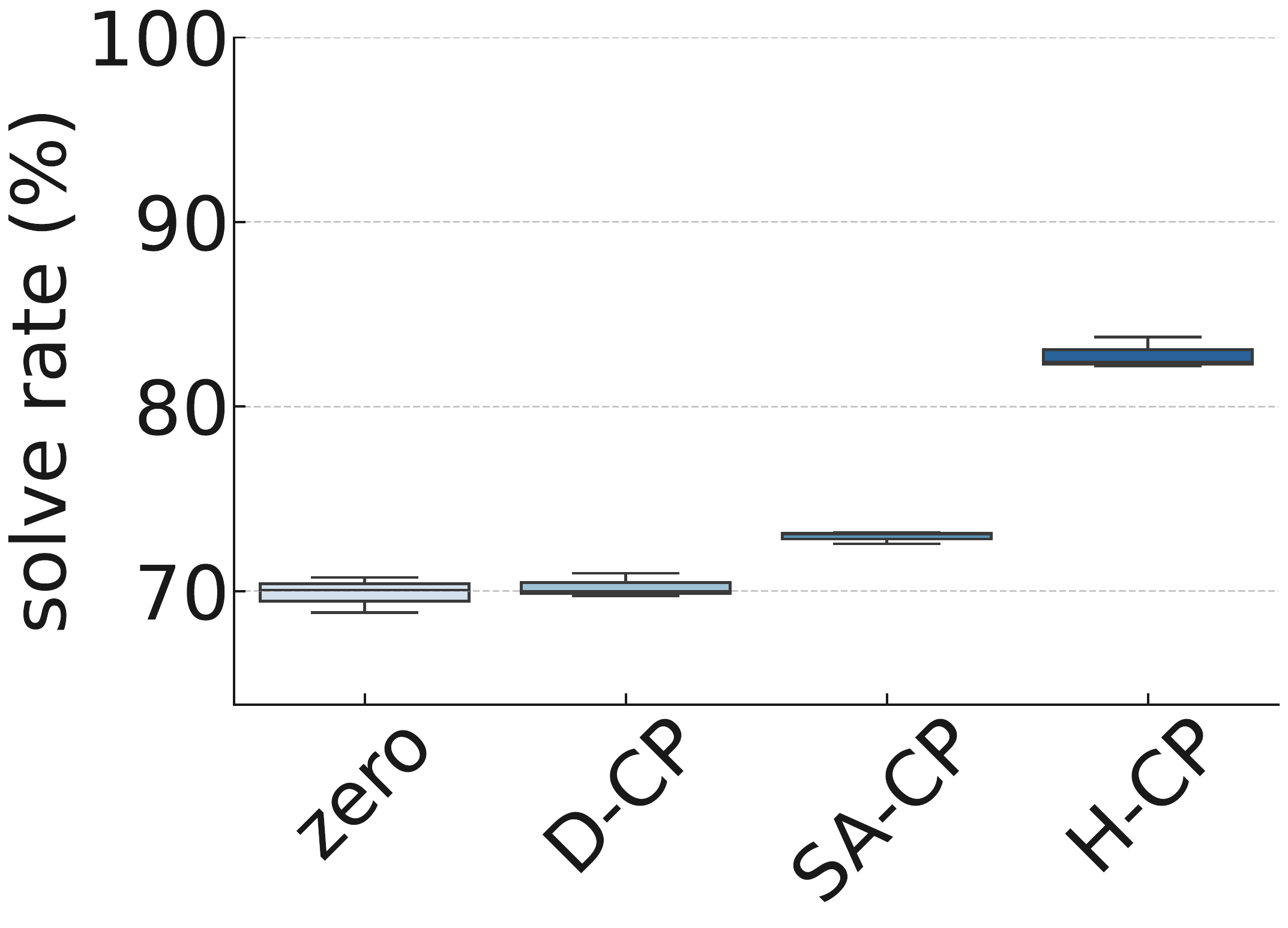}
    }
    \hfill
    \subfloat[Gemma 9B\label{fig:gemma_9b}]{
        \includegraphics[width=0.31\textwidth]{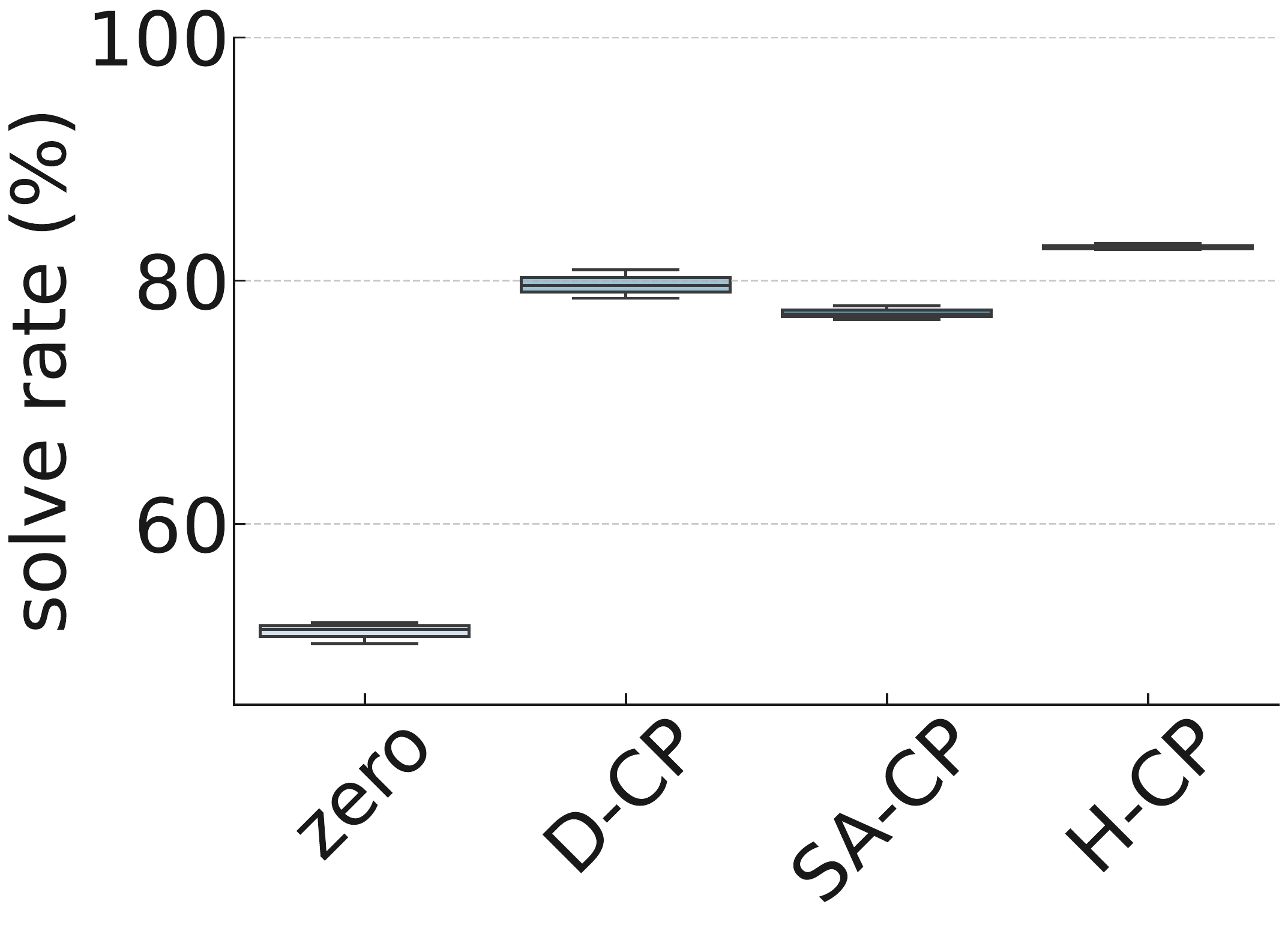}
    }
    \hfill
    \subfloat[Qwen 7B\label{fig:qwen_7b}]{
        \includegraphics[width=0.31\textwidth]{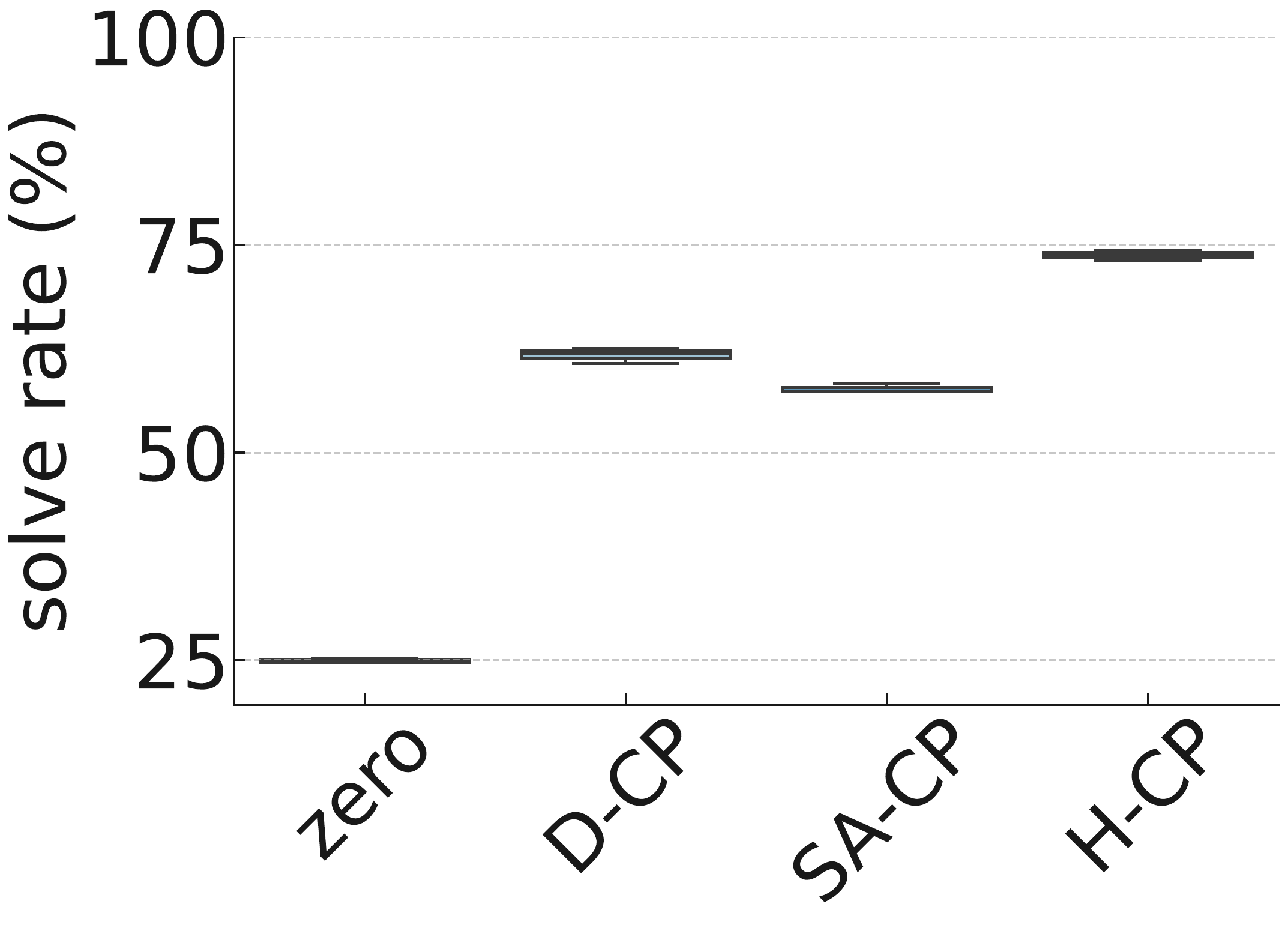}
    }
    \caption{Solve rates of CP strategies using mid-size models on GSM8k (3 repetitions).}
    \label{fig:llama}
\end{figure}

Figure \ref{fig:llama} presents the experimental results, comparing standard zero-shot prompting with D-CP, SA-CP, and H-CP (based on the self-adaptive prompt) for the mid-size model variants, i.e., LLaMA 8B, Gemma 9B, and Qwen 7B. CP variants consistently outperform zero-shot prompting across all models, demonstrating significant improvements.

\paragraph{Large Models.}

\begin{figure}[h!]
    \centering
    \subfloat[LLaMA 70B\label{fig:llama_70b}]{
        \includegraphics[width=0.31\textwidth]{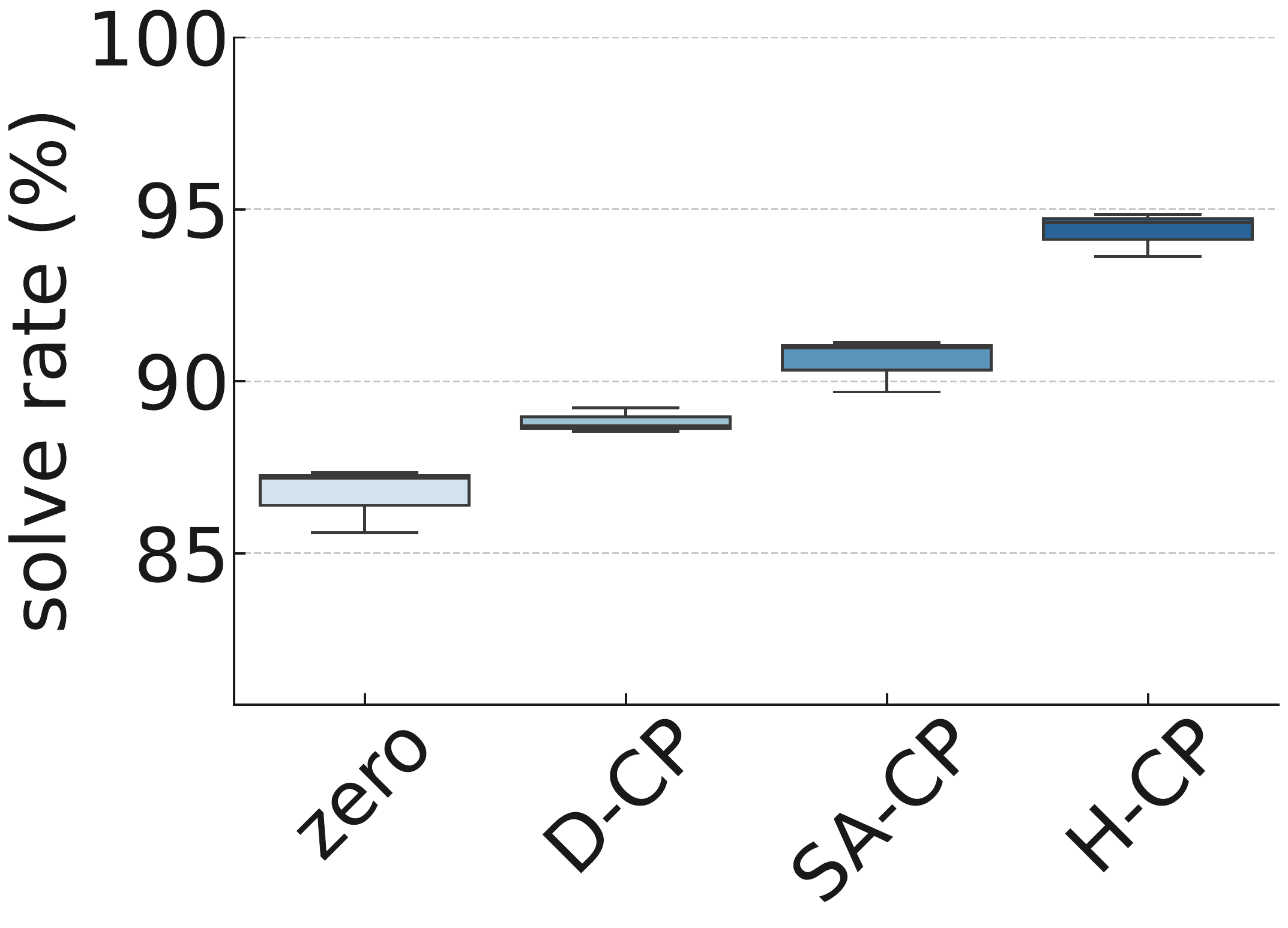}
    }
    \hfill
    \subfloat[Gemma 27B\label{fig:gemma_27b}]{
        \includegraphics[width=0.31\textwidth]{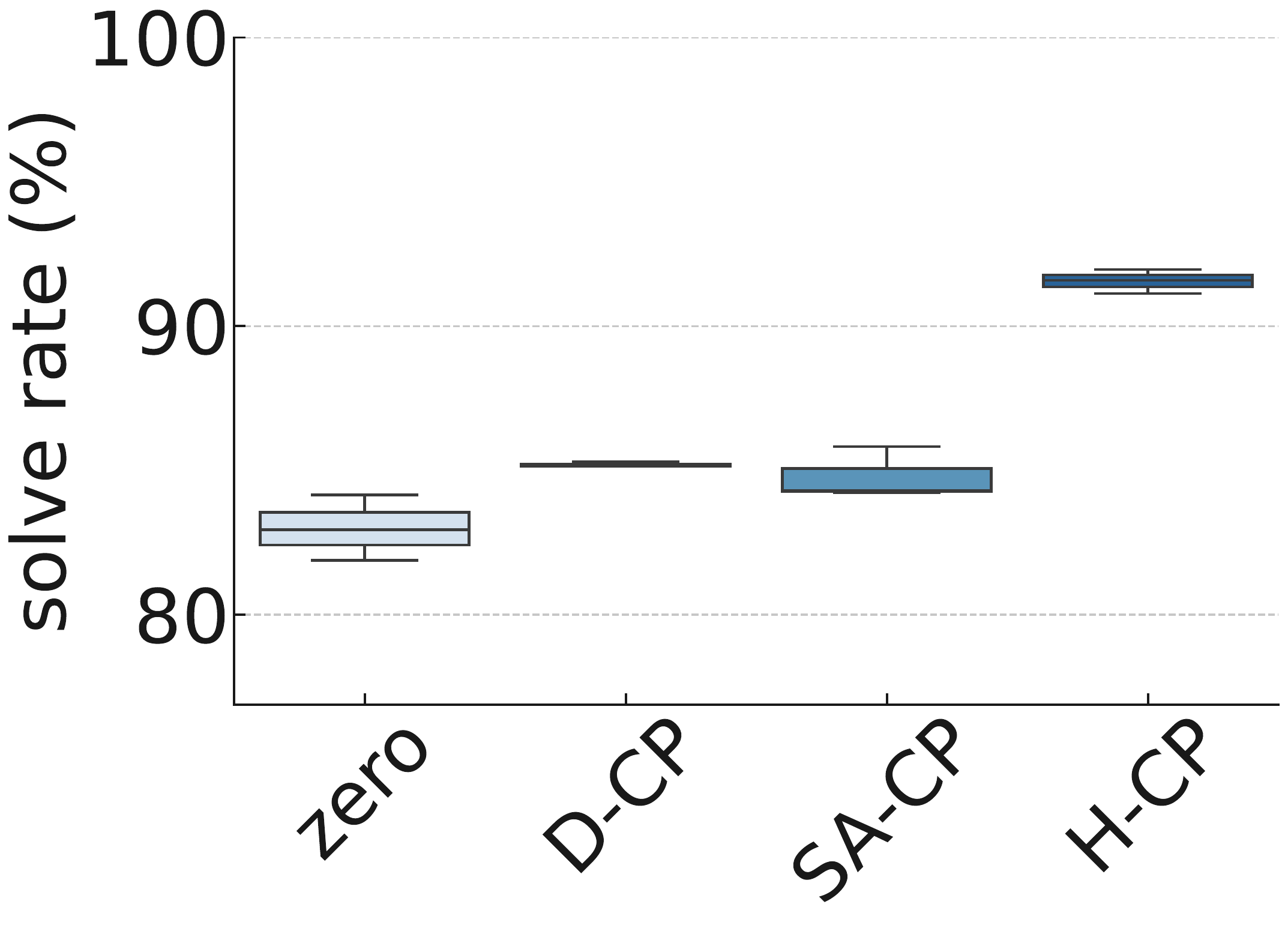}
    }
    \hfill
    \subfloat[Qwen 32B\label{fig:gemma_27}]{
        \includegraphics[width=0.31\textwidth]{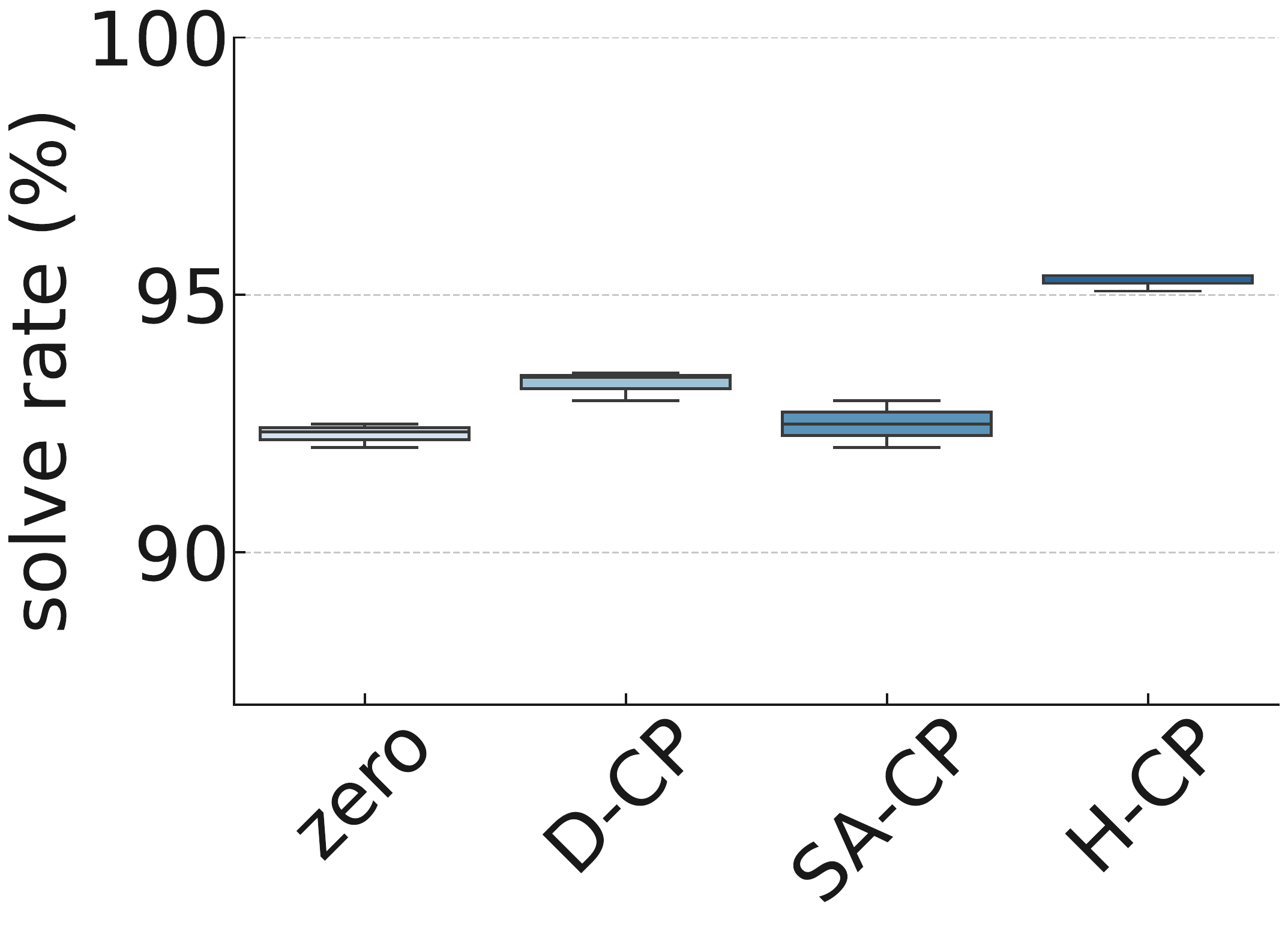}
    }
    \caption{Solve rates of CP strategies using large models on GSM8k (3 repetitions).}
    \label{fig:gemma}
\end{figure}

Figure \ref{fig:gemma} compares all variants across large models, including LLaMA 70B, Gemma 27B, and Qwen 32B, highlighting consistent improvements with CP. Notably, H-CP demonstrates a significant performance advantage, achieving an impressive 95\% solve rate on the LLaMA 70B model. While Qwen 32B delivers excellent results even with zero-shot prompting, its performance is further enhanced by CP, particularly with the hybrid CP variant.

\begin{wrapfigure}{r}{0.5\textwidth} % 'r' for right, 0.31\textwidth for width
    \vspace{-0.6cm}
    \centering
    \includegraphics[width=1.\textwidth]{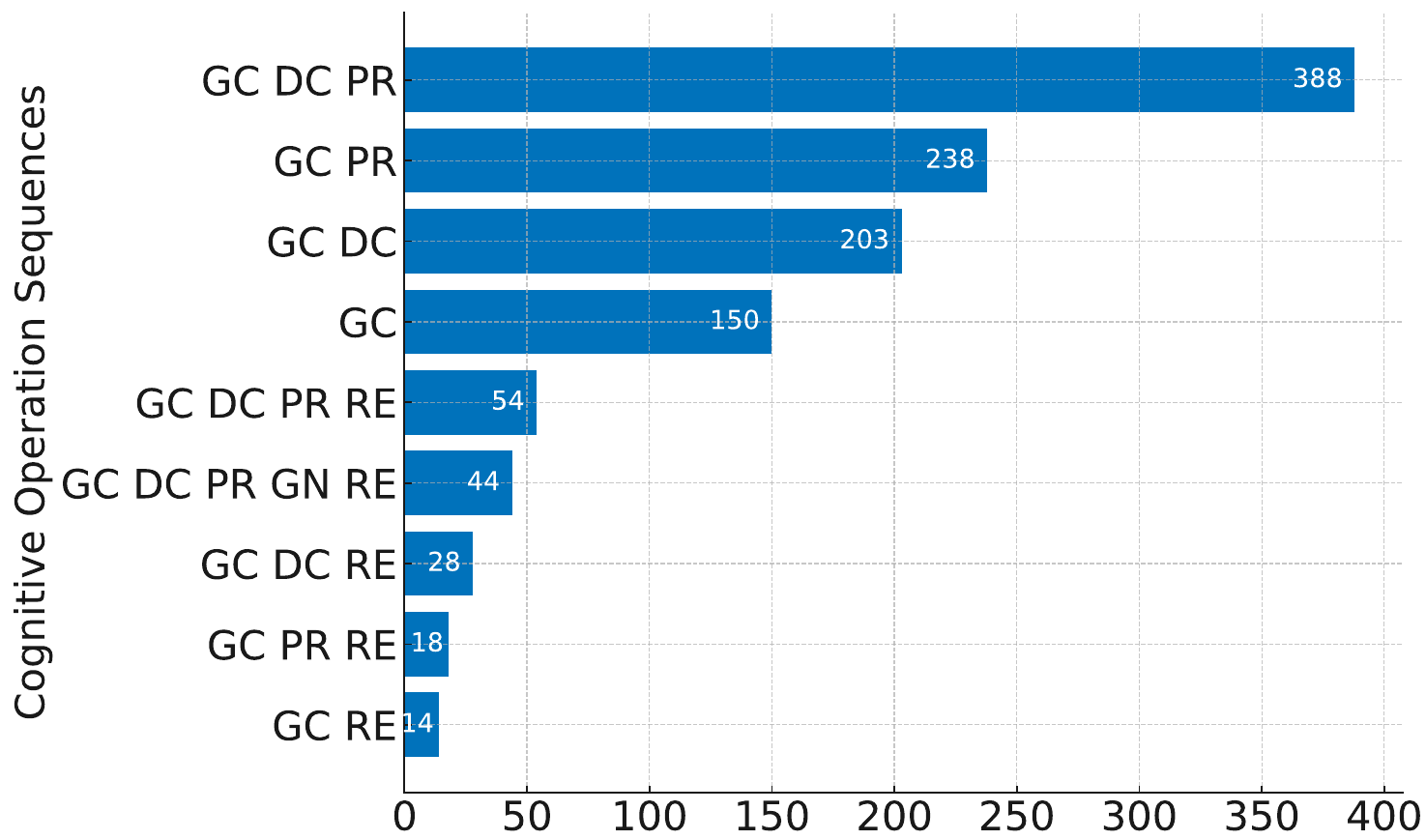}
    \caption{SA-CP sequences, LLaMA 70B.}
     \vspace{-0.3cm}
    \label{fig:sa-cops}
\end{wrapfigure}
Figure \ref{fig:sa-cops} shows the most frequent COP sequences\footnote{Goal clarification (GC), decomposition (DC), pattern recognition (PR), generalization (GN), reorganization (RE)} that have automatically been chosen by SA-CP on LLaMA 70B. GC-DC-PR is the most frequent sequence, indicating its fundamental role. Shorter sequences dominate, while longer, more complex sequences are used less often. We observed similar results for the other LLMs.

\section{Conclusions}
\label{sec:cons}

CP models human reasoning as a sequence of COPs delivered through structured prompts, fostering structured thinking through general or domain-specific COPs. Unlike example-based approaches like CoT, CP emphasizes high-level reasoning, making it highly adaptable across diverse tasks. %By specializing COPs for specific domains, CP effectively addresses a wide range of problem types.
Our experiments show that self-adaptive CP significantly boosts LLM performance on complex tasks, such as GSM8K math problems, with notable improvements for mid-size and larger models, though the proportional gain is greater for mid-size models. Additionally, the hybrid approach combining CoT few-shot prompting and CP delivers the best overall results across all experiments.

Our future work will focus on extending CP to additional domains and models, such as legal reasoning and strategic planning, to further validate its robustness in specialized tasks.

\bibliographystyle{unsrt}
\bibliography{literature}

\begin{thebibliography}{1}

\bibitem{brown2020language}
T.~B. Brown, B.~Mann, N.~Ryder, M.~Subbiah, J.~Kaplan, P.~Dhariwal,
  A.~Neelakantan, P.~Shyam, G.~Sastry, A.~Askell, S.~Agarwal, A.~Herbert-Voss,
  G.~Krueger, T.~Henighan, R.~Child, A.~Ramesh, D.~M. Ziegler, J.~Wu,
  C.~Winter, C.~Hesse, M.~Chen, E.~Sigler, M.~Litwin, S.~Gray, B.~Chess,
  J.~Clark, C.~Berner, S.~McCandlish, A.~Radford, I.~Sutskever, and D.~Amodei.
\newblock Language models are few-shot learners.
\newblock In {\em Neural Information Processing Systems (NeurIPS)}, volume~35,
  pages 24824--24837, 2022.

\bibitem{gsm8k}
K.~Cobbe, V.~Kosaraju, M.~Bavarian, M.~Chen, H.~Jun, L.~Kaiser, M.~Plappert,
  J.~Tworek, J.~Hilton, R.~Nakano, C.~Hesse, and J.~Schulman.
\newblock Training verifiers to solve math word problems.
\newblock {\em arXiv preprint arXiv:2110.14168}, 2021.

\bibitem{promptbreeder2022}
C.~Fernando, D.~Banarse, H.~Michalewski, S.~Osindero, and T.~Rocktäschel.
\newblock Promptbreeder: Self-referential self-improvement via prompt
  evolution.
\newblock {\em Neural Information Processing Systems (NeurIPS) Workshop}, 2023.

\bibitem{wei2022chain}
J.~Wei, X.~Wang, D.~Schuurmans, M.~Bosma, B.~Ichter, F.~Xia, E.~H. Chi, Q.~V.
  Le, and D.~Zhou.
\newblock Chain-of-thought prompting elicits reasoning in large language
  models.
\newblock In S.~Koyejo, S.~Mohamed, A.~Agarwal, D.~Belgrave, K.~Cho, and A.~Oh,
  editors, {\em Neural Information Processing Systems (NeurIPS) Workshop},
  volume~35, pages 24824--24837, 2022.

\bibitem{opro}
C.~Yang, X.~Wang, Y.~Lu, H.~Liu, Q.~V. Le, D.~Zhou, and X.~Chen.
\newblock Large language models as optimizers.
\newblock In {\em International Conference on Learning Representations (ICLR)},
  2024.

\bibitem{tree}
S.~Yao, D.~Yu, J.~Zhao, I.~Shafran, T.~Griffiths, Y.~Cao, and K.~Narasimhan.
\newblock Tree of thoughts: Deliberate problem solving with large language
  models.
\newblock In {\em Neural Information Processing Systems (NeurIPS)}, volume~36,
  pages 11809--11822, 2023.

\bibitem{yao2022react}
S.~Yao, J.~Zhao, D.~Yu, N.~Du, I.~Shafran, K.~R. Narasimhan, and Y.~Cao.
\newblock React: Synergizing reasoning and acting in language models.
\newblock In {\em International Conference on Learning Representations (ICLR)},
  2023.

\bibitem{ape}
Y.~Zhou, A.~I. Muresanu, Z.~Han, K.~Paster, S.~Pitis, H.~Chan, and J.~Ba.
\newblock Large language models are human-level prompt engineers.
\newblock In {\em International Conference on Learning Representations (ICLR)},
  2023.

\end{thebibliography}

\end{document}